\documentclass{article}

\usepackage[final,nonatbib]{neurips_2021}

\usepackage[utf8]{inputenc} 
\usepackage[T1]{fontenc}    
\usepackage{hyperref}       
\usepackage{url}            
\usepackage{booktabs}       
\usepackage{amsfonts}       
\usepackage{nicefrac}       
\usepackage{microtype}      
\usepackage{xcolor}         
\usepackage{graphicx}

\usepackage{tabularx}
\usepackage{siunitx}
\usepackage{array}
\usepackage{multirow} 
\usepackage{multicol}
\usepackage{rotating}
\usepackage[export]{adjustbox}

\newcommand{\secref}[1]{Section~\ref{#1}}
\newcommand{\figref}[1]{Figure~\ref{#1}}
\newcommand{\tabref}[1]{Table~\ref{#1}}

\newcommand{\rot}[1]{\rotatebox[origin=c]{90}{#1}}
\newcolumntype{P}[1]{>{\centering\arraybackslash}p{#1}}
\title{7\textsuperscript{th} AI Driving Olympics: \\ 1\textsuperscript{st} Place Report for Panoptic Tracking}

\author{
  Rohit Mohan \\
  Department of Computer Science\\
  University of Freiburg\\
  \texttt{mohan@cs.uni-freiburg.de} \\
  \And
  Abhinav Valada \\
  Department of Computer Science\\
  University of Freiburg\\
  \texttt{valada@cs.uni-freiburg.de} \\
}

\begin{document}

\maketitle

\begin{abstract}
In this technical report, we describe our \textit{EfficientLPT} architecture that won the panoptic tracking challenge in the 7\textsuperscript{th} AI Driving Olympics at NeurIPS 2021. Our architecture builds upon the top-down EfficientLPS panoptic segmentation approach. EfficientLPT consists of a shared backbone with a modified \mbox{EfficientNet-B5} model comprising the proximity convolution module as the encoder followed by the range-aware FPN to aggregate semantically rich range-aware multi-scale features. Subsequently, we employ two task-specific heads, the scale-invariant semantic head and hybrid task cascade with feedback from the semantic head as the instance head. Further, we employ a novel panoptic fusion module to adaptively fuse logits from each of the heads to yield the panoptic tracking output. Our approach exploits three consecutive accumulated scans to predict locally consistent panoptic tracking IDs and  also the overlap between the scans to predict globally consistent panoptic tracking IDs for a given sequence. The benchmarking results from the 7\textsuperscript{th} AI Driving Olympics at NeurIPS 2021 show that our model is ranked \#1 for the panoptic tracking task on the Panoptic nuScenes dataset.  
\end{abstract}

\section{Introduction}
\label{sec:intro}

Holistic comprehension of dynamic scenes is a crucial prerequisite for autonomous robots to navigate in the environment. Providing robots with this ability allows them to perceive and reason about the elements in the scene, their actions~\cite{zurn2020self}, and the occurring events~\cite{younes2021catch}. This is an essential step for robots to make complex decisions. In this direction, different works address various tasks in the autonomy pipeline such as dynamic object segmentation~\cite{bevsic2020dynamic}, tracking~\cite{valverde2021there}, and behavior prediction~\cite{radwan2020multimodal}. The advance towards a more detailed scene representation can be seen in the most recently proposed tasks such as birds-eye-view panoptic segmentation~\cite{gosala2021bird} and panoptic tracking~\cite{hurtado2020mopt}. LiDAR panoptic segmentation allows to reason about all the elements in the scene, classifying them into \textit{stuff} and \textit{thing} categories. \textit{Stuff} classes are defined as amorphous regions such as roads and buildings, while \textit{thing} classes are defined as object instances such as people, cars, and cyclists. In this task, the main goal is to assign a semantic class ID to a pixel of an image if it belongs to \textit{stuff} classes and assigns a class label as well as an instance ID if the pixel belongs to a \textit{thing} class. Although panoptic segmentation allows for dense classification of the scene while simultaneously identifying each instance, it is performed in a frame-wise manner. Therefore, this task is not able to represent the temporal dynamics of the scene inherent to many real-world applications. To address this problem, more recent tasks aim to represent this temporal information with the correspondence between different frames at the object level in LiDAR panoptic tracking~\cite{hurtado2020mopt}.

Considering this, we address the panoptic tracking problem with our Efficient LiDAR 
Panoptic Tracking (EfficientLPT) architecture, a two-module approach that follows the clip-match paradigm. It comprises an architecture that processes accumulated scans from the current timestep $t$ with the past two scans from $t-1$ and $t-2$ to compute the panoptic segmentation output. The aggregated panoptic output when segregated to the corresponding $t$, $t-1$, and $t-2$ results in panoptic predictions with temporally consistent instance IDs. We refer to these aforementioned IDs as locally consistent panoptic tracking IDs as the consistency exists only for any three consecutive scans. We obtain globally consistent panoptic tracking IDs with point intersections between the consecutive overlapping scan points. The panoptic segmentation approach of our architecture is based on the EfficientLPS~\cite{sirohi2021efficientlps} model that comprises a novel shared backbone that encodes with strengthened geometric transformation modeling capacity and aggregates semantically rich range-aware multi-scale features. Further, it incorporates scale-invariant semantic and instance segmentation heads along with a panoptic fusion module to yield the final output. 

We evaluate our architecture on the Panoptic nuScenes~\cite{fong2021panoptic} dataset and benchmark on the panoptic tracking challenge in the 7\textsuperscript{th} AI Driving Olympics at NeurIPS 2021. The goal of the challenge is to encourage the development of LiDAR perception systems that are robust~\cite{bevsic2021unsupervised}. The panoptic tracking track of the challenge requires participants to predict the semantic categories of \textit{stuff} and \textit{thing} classes along with temporally-consistent IDs for each \textit{thing} object in the sequence. The approaches submitted to the competition are ranked using the PAT metric~\cite{fong2021panoptic}. Our approach obtains a PAT score of $70.4\%$ achieving the \#1 position in the leaderboard of the panoptic tracking challenge. 

\section{Methodology}
\label{sec:method}

\begin{figure*}
    \centering
    \includegraphics[width=\linewidth]{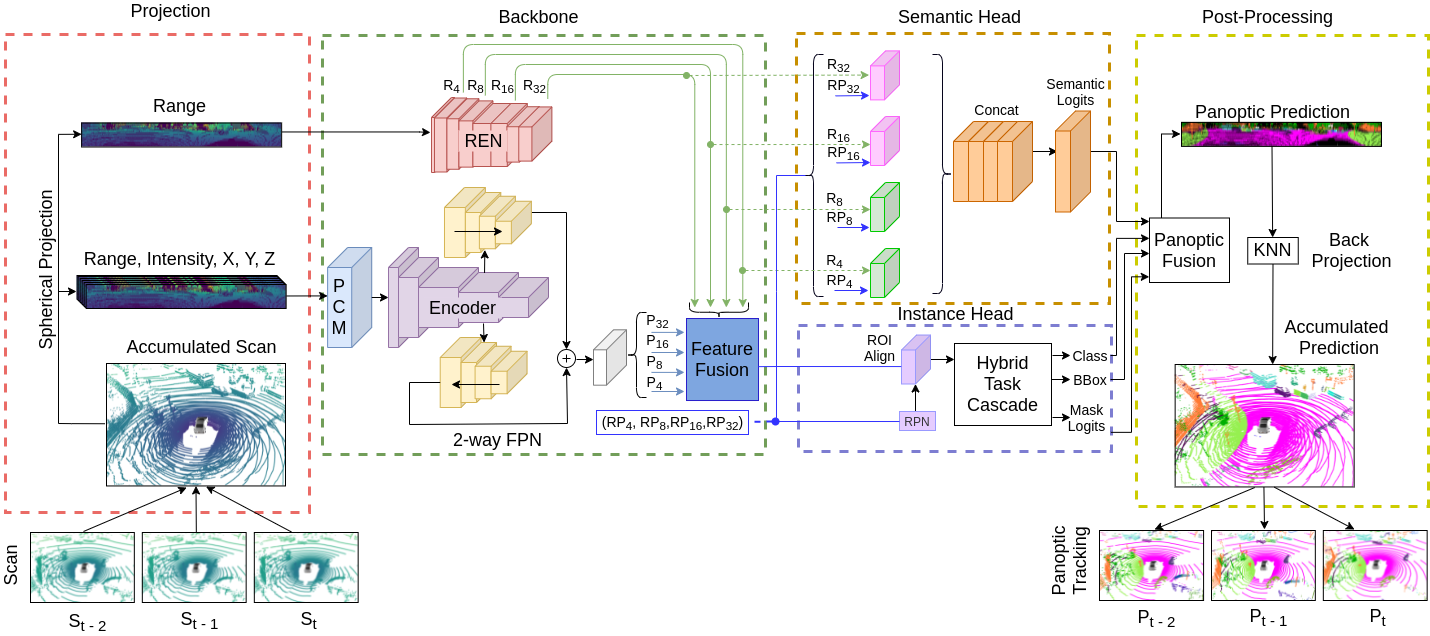}
    \caption{Illustration of our proposed EfficientLPT architecture for LiDAR panoptic tracking. The point clouds are first accumulated for three consecutive scans, followed by projecting it into the 2D domain using spherical projection and fed as an input to the Proximity Convolution Module (PCM). Subsequently, we employ a modified EfficientNet model with the 2-way FPN and the Range Encoder Network (REN) in parallel. The fusion of outputs from REN and 2-way FPN results in the overall range-aware FPN topology. The output of the range-aware FPN is then fed to the semantic and instance heads. Following, logits of each head are then combined in the panoptic fusion module. Next, the output of this module is projected back into the 3D domain using KNNs. Finally, the accumulated panoptic segmentation is segregated to obtain the panoptic tracking predictions.}
    \label{fig:efficient_lpt}
\end{figure*}

We propose the EfficientLPT architecture that builds upon our top-down LiDAR panoptic segmentation approach EfficentLPS. \figref{fig:efficient_lpt} illustrates the topology of our EfficientLPT architecture. First, we merge the point clouds from the LiDAR scanner at $t$, $t-1$ and $t-2$ timesteps where $t$ is the current scan time. We then project the point cloud using spherical projection. The projected representation consists of five channels that are range, intensity, and the (x,y,z) coordinates. Subsequently, we employ a shared backbone which consists of modified EfficientNet-B5 as an encoder with a Proximity Convolution Module (PCM) as its initial layer that aids in modeling geometric transformations. Following, we employ the 2-way FPN~\cite{mohan2020efficientps} on top of the encoder. Parallelly, we employ the Range Encoder Network (REN) similar to EfficientLPS which takes the range channel of the projected accumulated point cloud as input. We then fuse the multi-scale outputs of the encoder and REN to obtain the range-aware feature pyramid that enhances the ability to distinguish adjacent objects at different distances, referred to as Range-Aware FPN. Subsequently, we employ parallel semantic segmentation and instance segmentation heads. The semantic head utilizes range-guided depth-wise atrous separable convolution in its associated modules, namely, Dense Prediction Cells(DPC)~\cite{chen2018searching} and Large Scale Feature Extractor (LSFE)~\cite{mohan2020efficientps} to enable capturing of scale-invariant long-range contextual features and fine features. 

Unlike EfficientLPS that employs Mask R-CNN~\cite{he2017mask} as its instance head, we opt for Hybrid Task Cascade~\cite{chen2019hybrid} (HTC) as our instance head. HTC performs cascaded refinement with semantic feature fusion resulting in vastly improved detection and segmentation of \textit{thing} instances. We then fuse the logits from both the heads with a parameter-free panoptic fusion module~\cite{mohan2020efficientps} to yield the accumulated scans panoptic predictions in the projection domain. Next, we re-project the predictions into the 3D space using the KNN algorithm~\cite{milioto2019rangenet++} and segregate it into panoptic predictions for scans from timestep $t$, $t-1$ and $t-2$. Thus, obtaining the local panoptic tracking output. Lastly, we assign global panoptic track IDs that are consistent through the entirety of the given sequence. To do so, we begin with the prediction of local panoptic track IDs of the first trio of scans and preserve these panoptic track IDs for the object points that overlap with the next trio of scans. In the case of non-overlapping object points with IDs not associated with any overlapping points, we assign a new unique track ID. We reiterate the aforementioned steps until a given sequence is processed in its entirety to obtain the final panoptic tracking output of the sequence.  

\section{Training}
\label{sec:training}

In this section, we first describe the datasets in \secref{sec:dataset} and then present the training protocol that we employ in \secref{sec:protocol}. Moreover, we use the PyTorch~\cite{paszke2019pytorch} deep learning library for implementing our architecture and we trained our model on a system with an Intel Xenon@2.20GHz processor and NVIDIA RTX A6000 GPUs.

\paragraph{Panoptic nuScenes Dataset}
\label{sec:dataset}

We evaluate our approach for panoptic tracking on the recently introduced Panoptic nuScenes dataset~\cite{fong2021panoptic}. The dataset contains $16$ semantic labels for evaluation with temporally consistent instance IDs for \textit{thing} classes. It consists of $6$ \textit{stuff} classes and $10$ thing classes. Further, the dataset contains $1000$ scenes, out of which $700$ scenes are used for the training set, $150$ scenes for the validation set, and the rest $150$ scenes for the test set. 

\paragraph{Training Protocol}
\label{sec:protocol}

We train EfficientLPT on projected point clouds of $4096\times256$ resolution where we employ bilinear interpolation on the projections obtained from LiDAR scan points and nearest-neighbor interpolation on the groundtruth. We initialize the backbone of our architecture with weights from the EfficientNet model pre-trained on the ImageNet dataset and initialize the weights of the iABN sync layers to $1$. We use Xavier initialization~\cite{glorot2010understanding} for the other layers, zero constant initialization for the biases and we use Leaky ReLU with a slope of $0.01$. We use the same hyperparameters as~\cite{sirohi2021efficientlps} unless explicitly mentioned in this report. We train our model with Stochastic Gradient Descent(SGD) with a momentum of $0.9$ using a multi-step learning rate schedule i.e. we start with an initial base learning rate and train the model for a certain number of iterations, followed by lowering the learning rate by a factor of $10$ at each milestone and continue training for $50,000$ iterations. We use an initial learning rate $lr_{base}$ of $0.01$ and successively reduce it by a factor of $10$ at $32,000$ and $44,000$ iterations. At the beginning of the training, we have a warm-up phase where the $lr_{base}$ is increased linearly from $\frac{1}{3}\cdot lr_{base}$ to $lr_{base}$ in $200$ iterations. We train our EfficientLPT with a batch size of 16 on 4 NVIDIA RTX A6000 GPUs. Please note that our benchmarked model is only trained on the training set i.e exclusive of the validation set.

\begin{table}
\centering
\caption{7\textsuperscript{th} AI Driving Olympics, NeurIPS 2021. Competition: Panoptic Tracking Track performance on the test set of the Panoptic nuScenes dataset. PAT:  Panoptic Tracking, PQ: Panoptic Quality, and TQ: Tracking Quality. All scores are in [$\%$].}
\label{tab:resultstab}
\begin{tabular}{p{5.0cm}p{1.5cm}p{1.5cm}p{1cm}p{1cm}p{1cm}}
\toprule
Methods & PAT & PQ  & TQ & PTQ & LSTQ \\
\midrule
PanopticTrackNet~\cite{hurtado2020mopt} & $45.7$ & $51.7$ & $40.9$ & $50.9$ & $43.4$\\ 
4D-PLS~\cite{aygun20214d} & $60.5$ & $62.0$ & $67.6$ & $55.6$ & $57.8$ \\
EfficientLPS~\cite{sirohi2021efficientlps} + Kalman Filter & $67.1$ & $63.6$ & $71.2$ & $62.3$ & $63.7$\\
\midrule
EfficientLPT (Ours) & \textbf{$70.4$} & \textbf{$67.9$} & \textbf{$73.1$} & \textbf{$67.0$}& \textbf{$66.0$} \\
\bottomrule
\end{tabular}
\end{table}

\section{Benchmark Results}
\label{sec:benchmark}

\begin{figure*}
\centering
\footnotesize
\setlength{\tabcolsep}{0.1cm}
{\renewcommand{\arraystretch}{0.5}
\begin{tabular}{P{0.2cm}P{4.4cm}P{4.4cm}P{4.4cm}}
& \raisebox{-0.4\height}{Scan\textsubscript{(t-2)}} & \raisebox{-0.4\height}{Scan\textsubscript{(t-1)}} & \raisebox{-0.4\height}{Scan\textsubscript{(t)}}\\
\\
\rot{(a)}  &\raisebox{-0.4\height}{\includegraphics[width=\linewidth,frame]{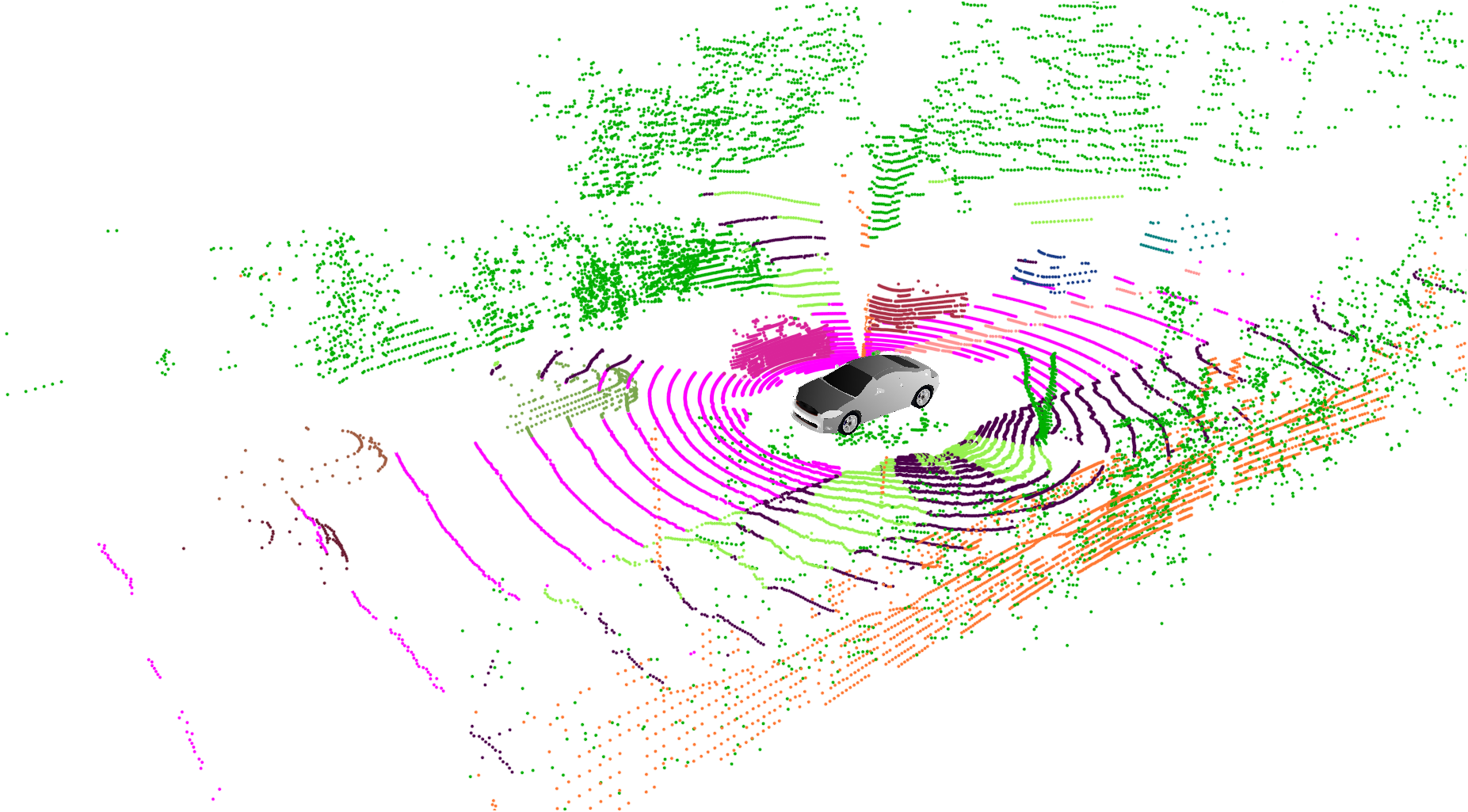}} & \raisebox{-0.4\height}{\includegraphics[width=\linewidth,frame]{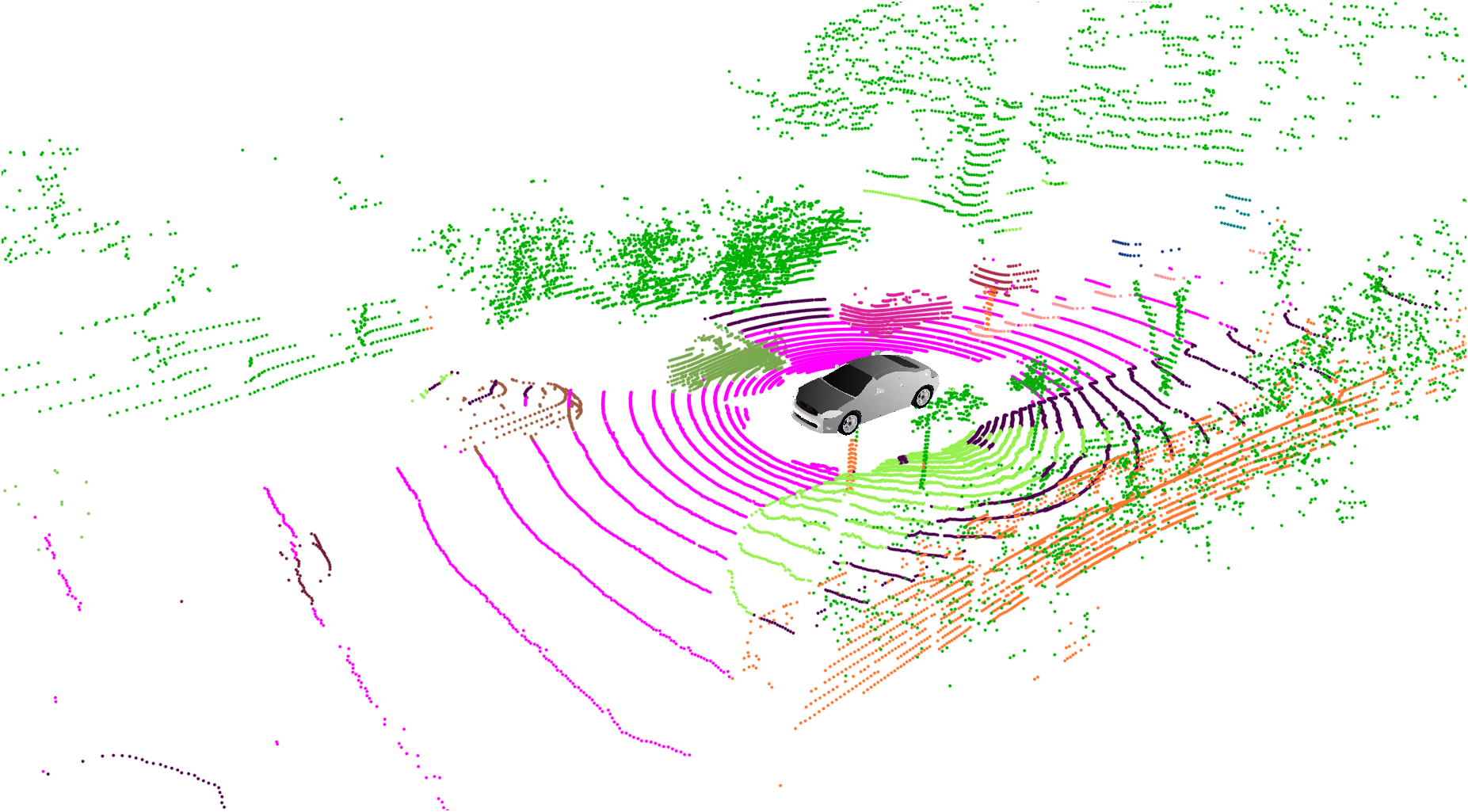}} & \raisebox{-0.4\height}{\includegraphics[width=\linewidth,frame]{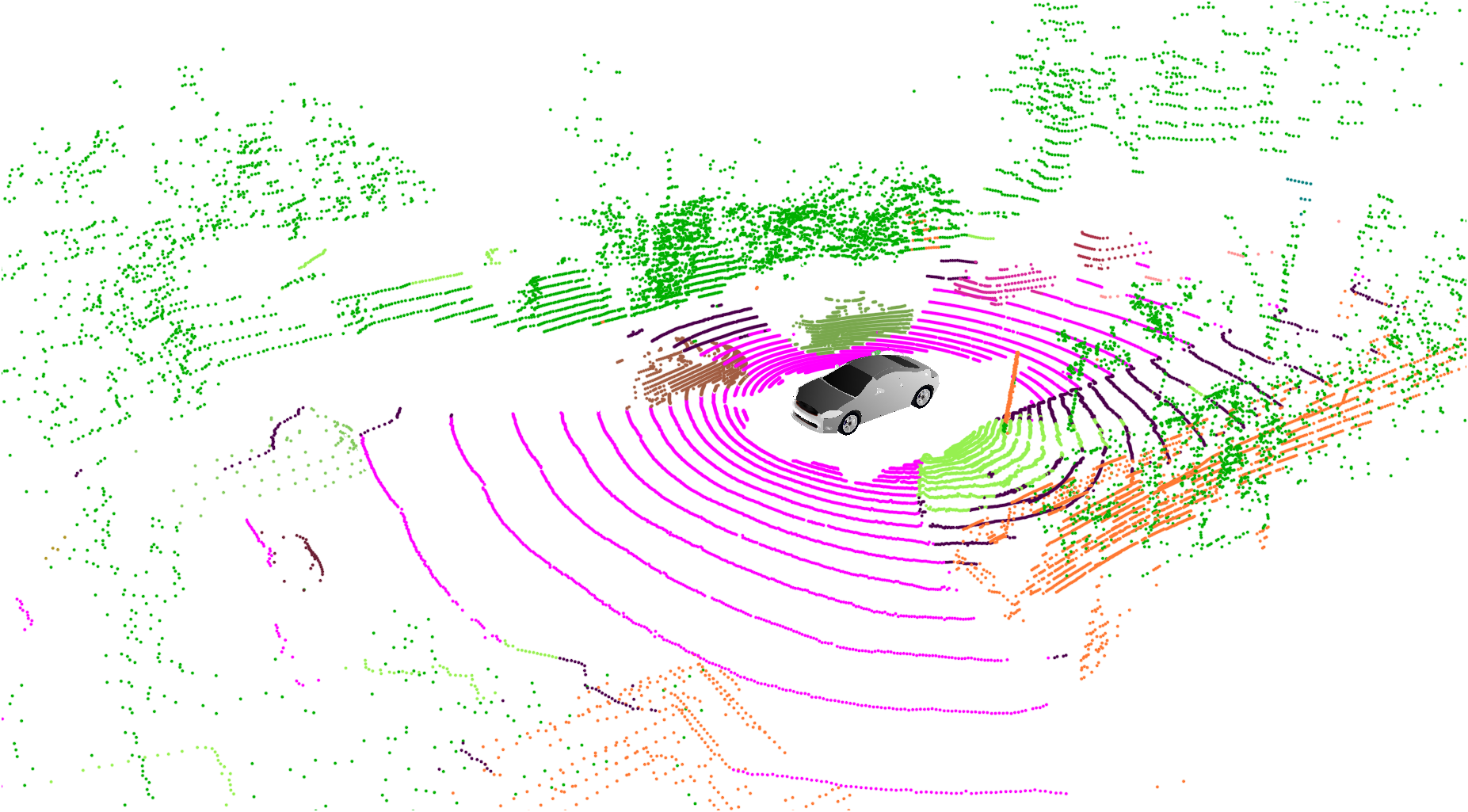}} \\
\\
\rot{(b)}  &\raisebox{-0.4\height}{\includegraphics[width=\linewidth,frame]{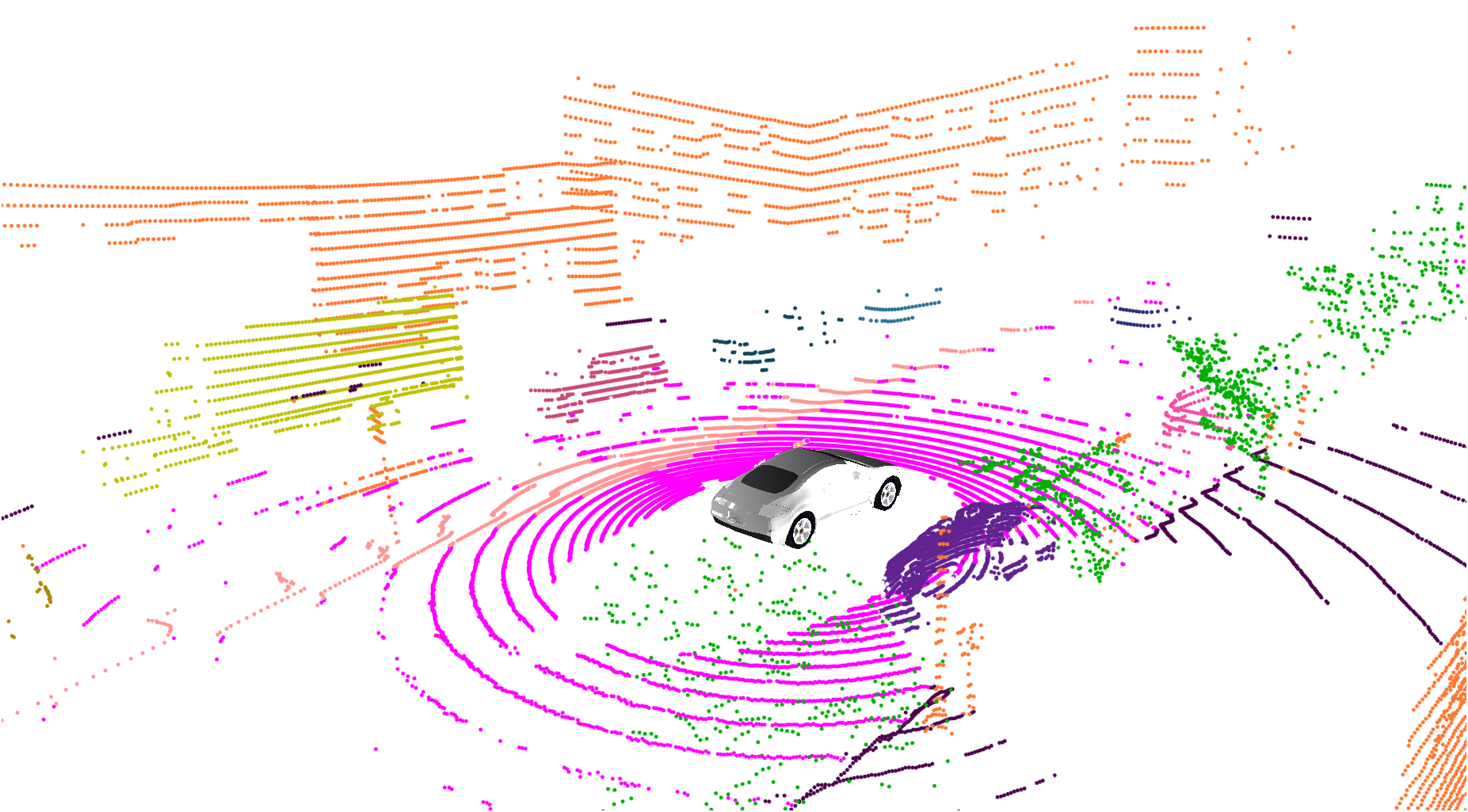}} & \raisebox{-0.4\height}{\includegraphics[width=\linewidth,frame]{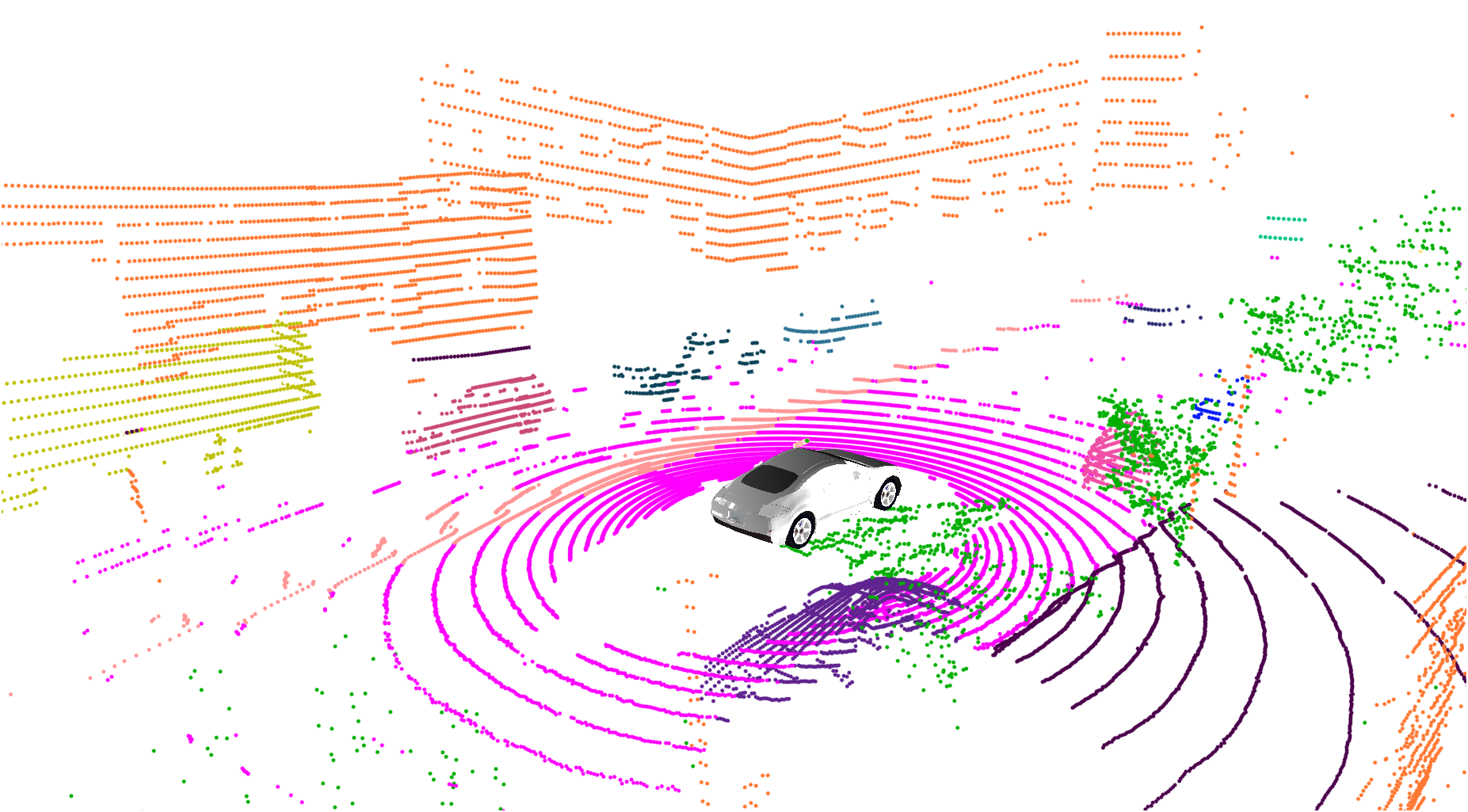}} & \raisebox{-0.4\height}{\includegraphics[width=\linewidth,frame]{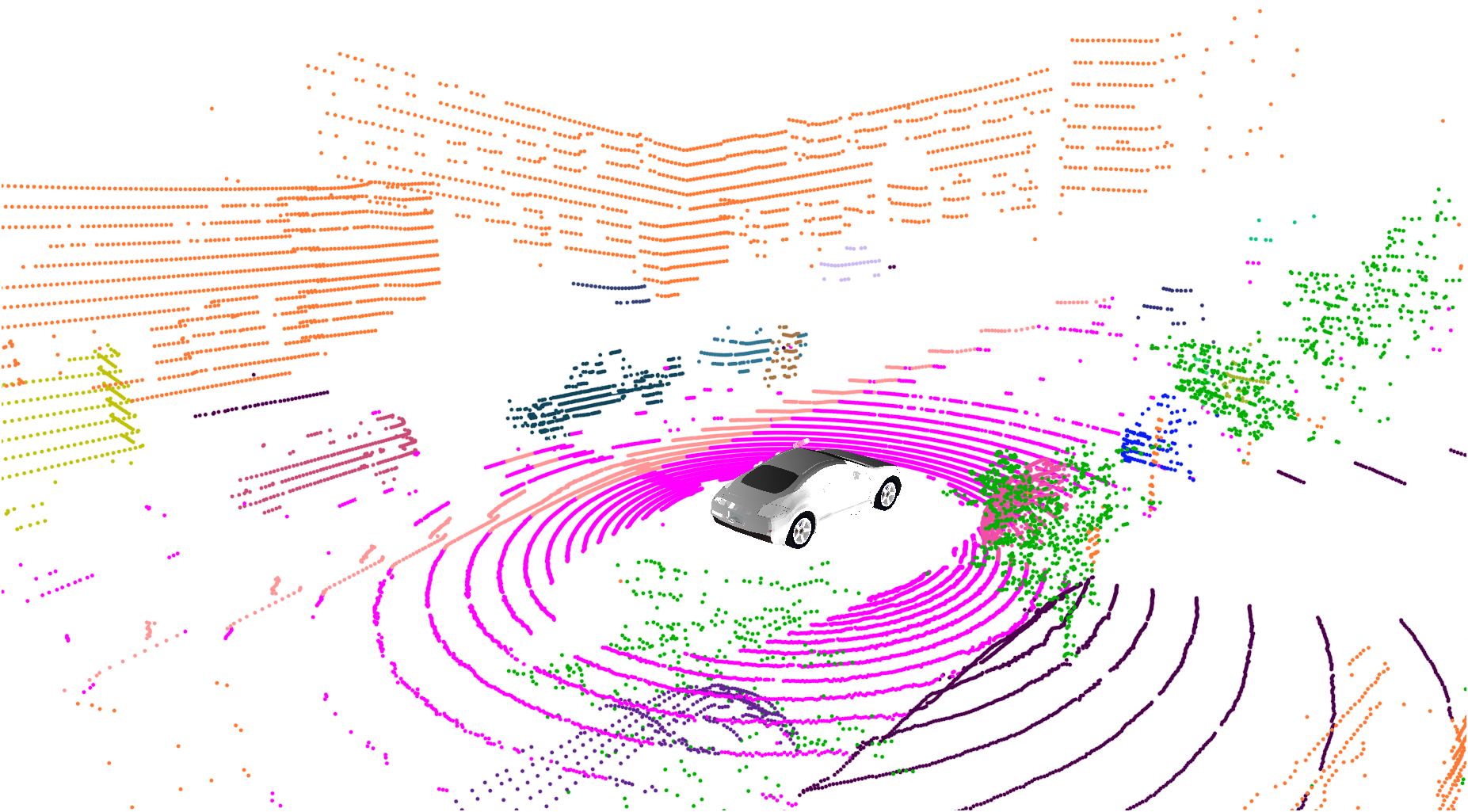}} \\
\\
\rot{(c)}  &\raisebox{-0.4\height}{\includegraphics[width=\linewidth,frame]{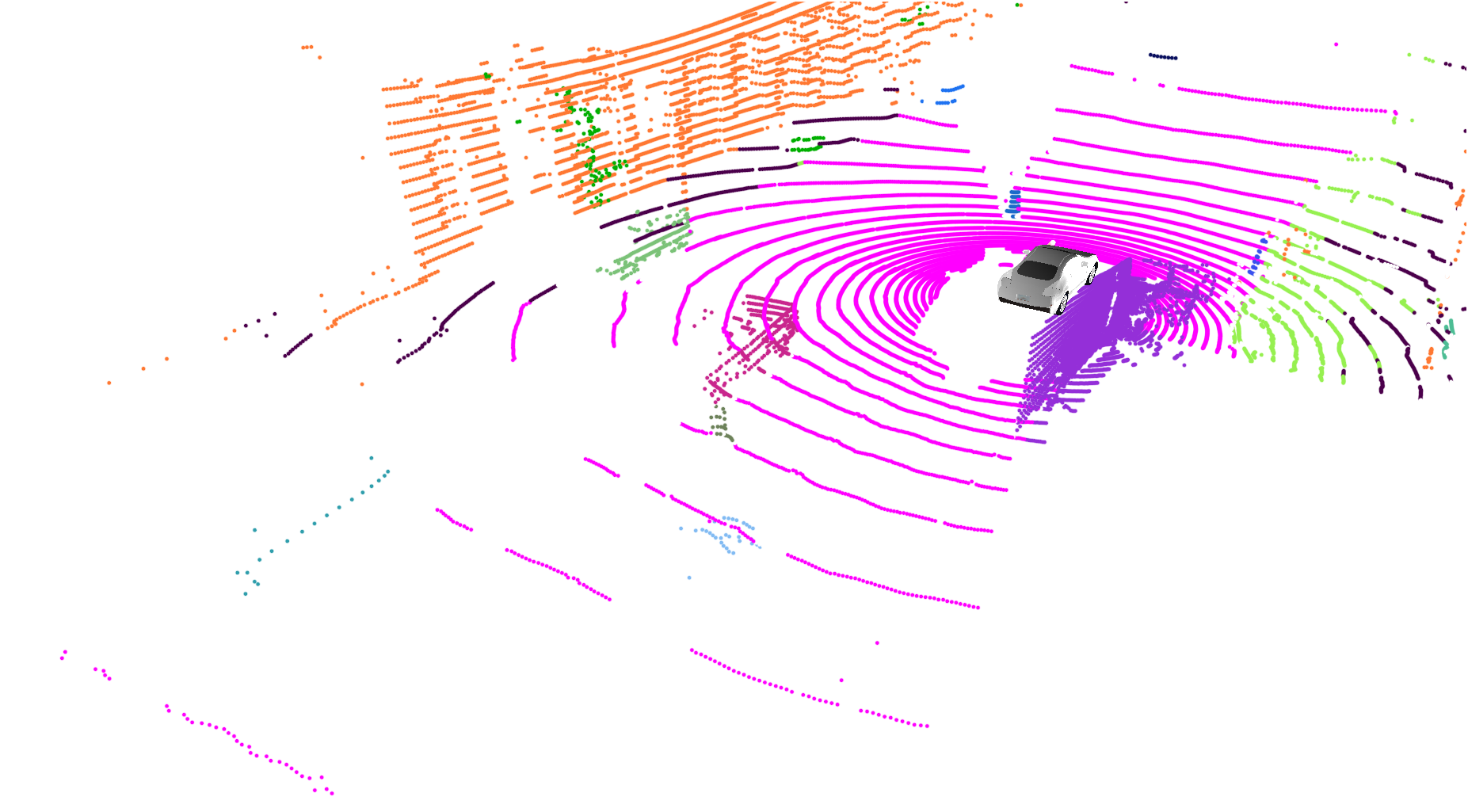}} & \raisebox{-0.4\height}{\includegraphics[width=\linewidth,frame]{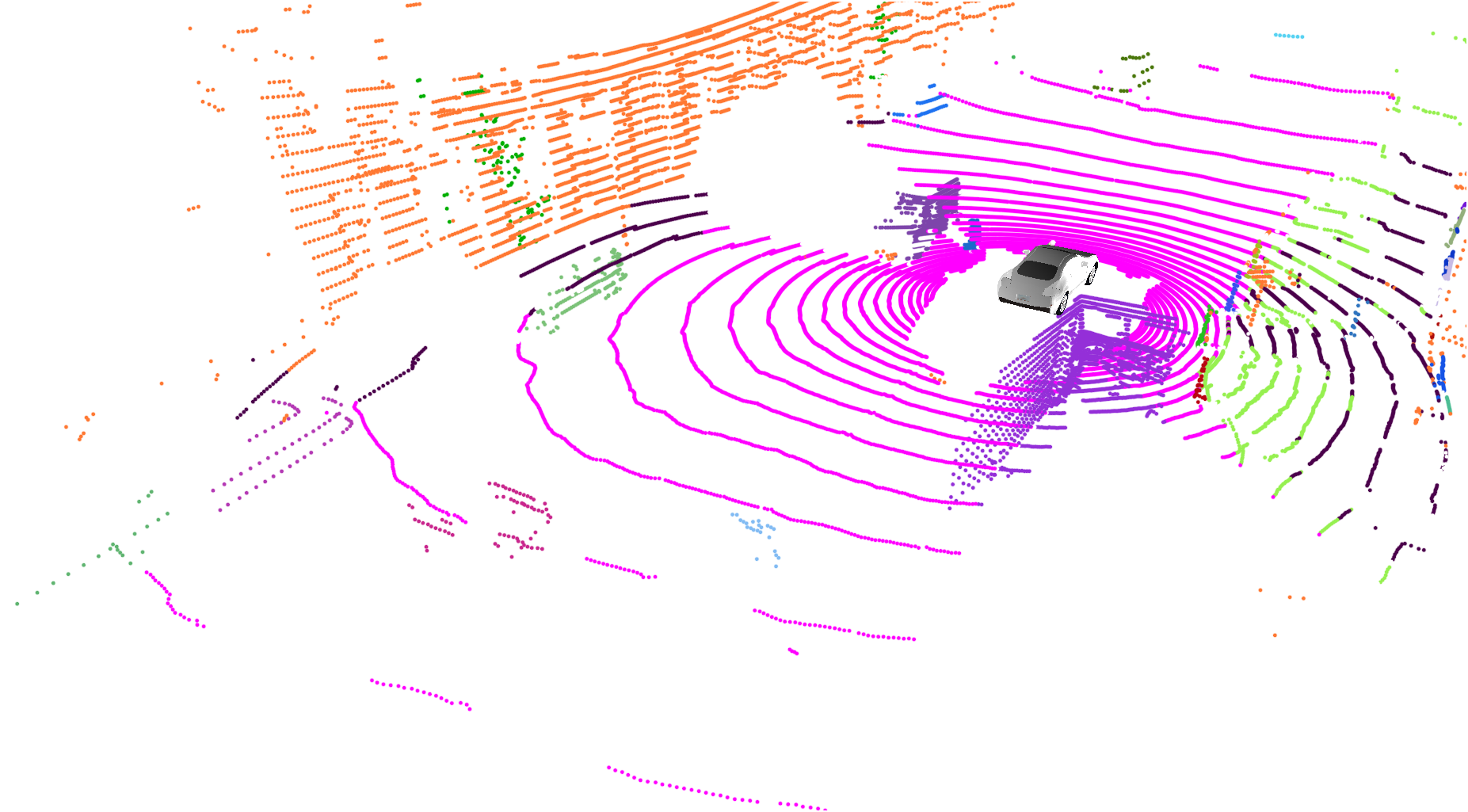}} & \raisebox{-0.4\height}{\includegraphics[width=\linewidth,frame]{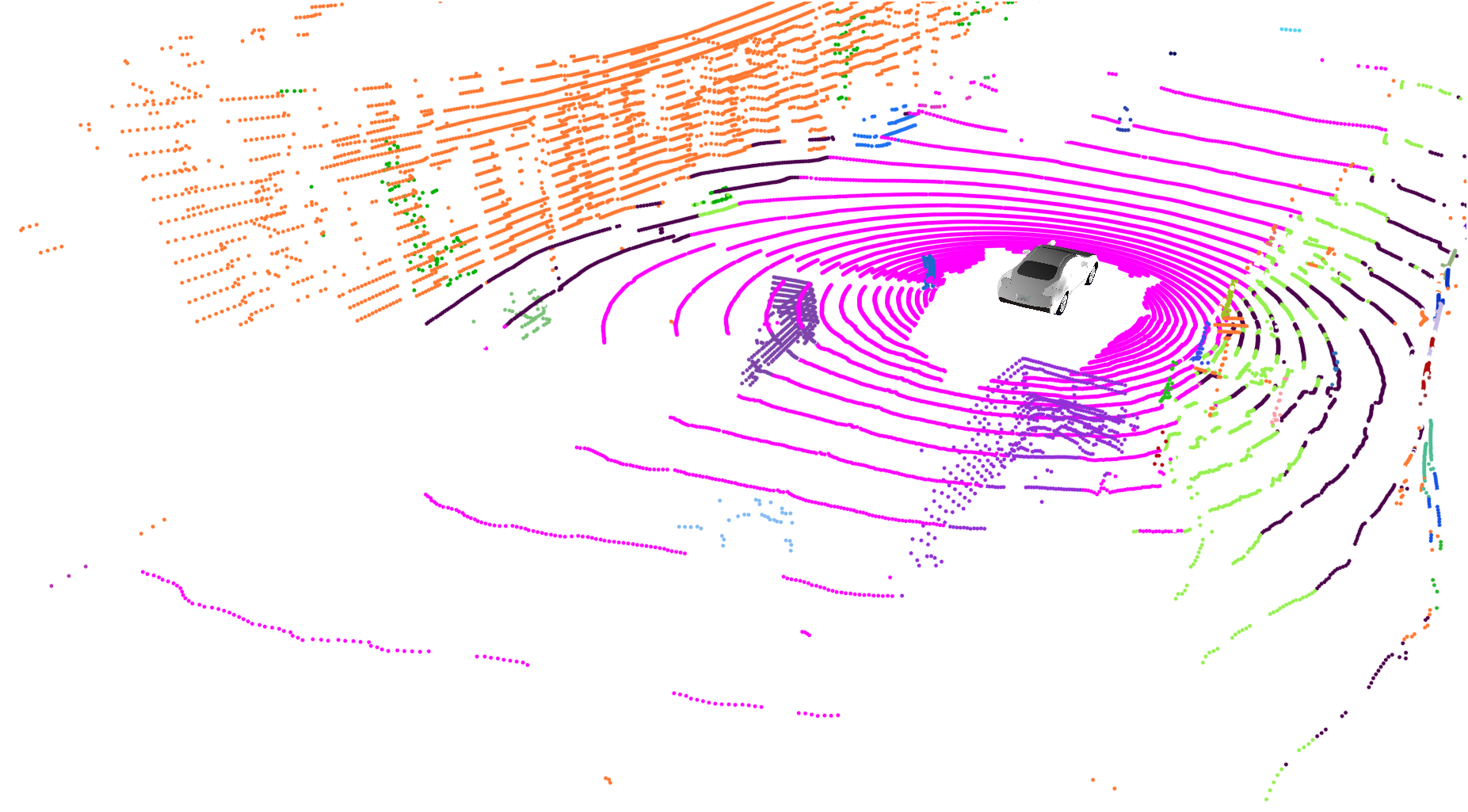}} \\
\\
\rot{(d)}  &\raisebox{-0.4\height}{\includegraphics[width=\linewidth,frame]{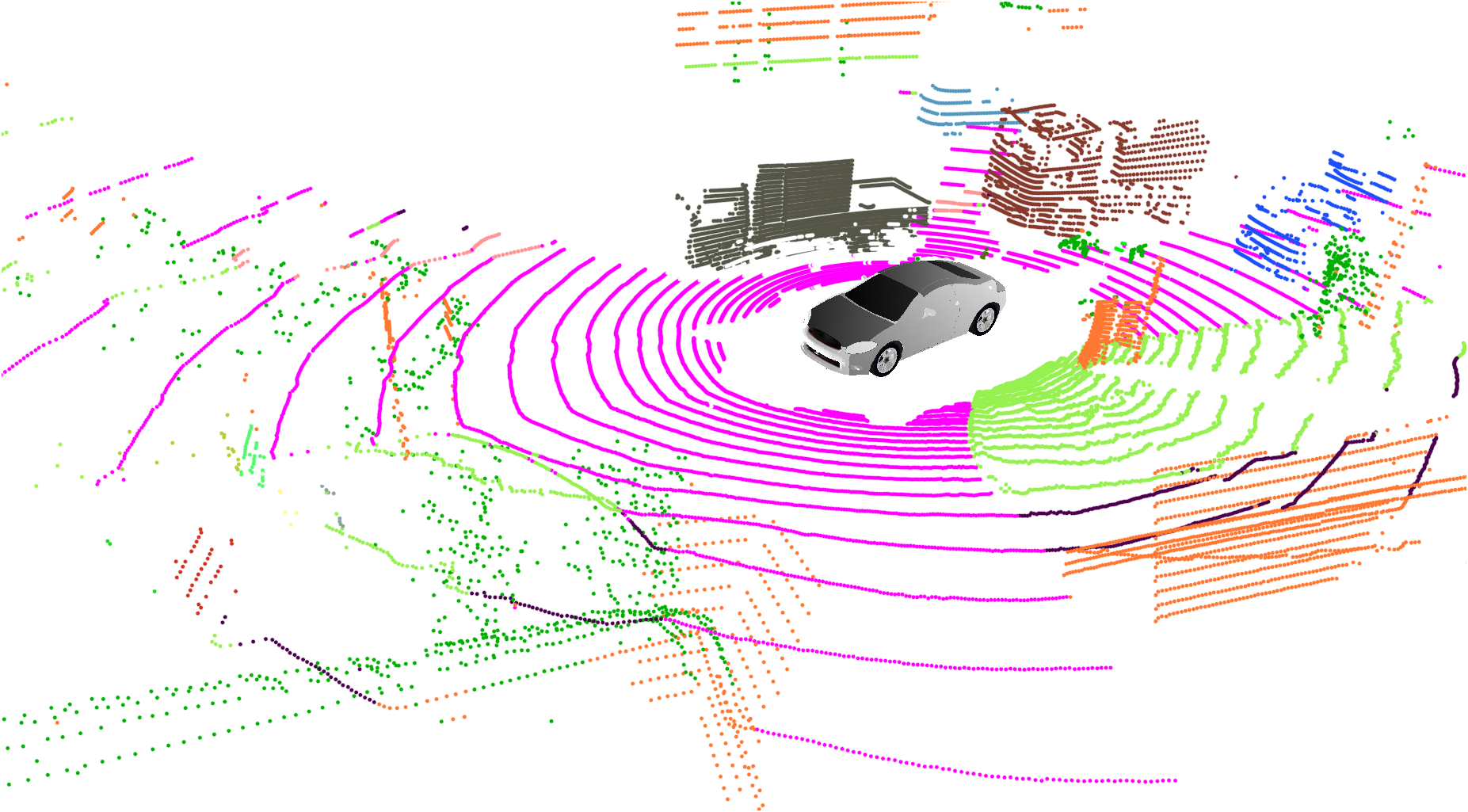}} & \raisebox{-0.4\height}{\includegraphics[width=\linewidth,frame]{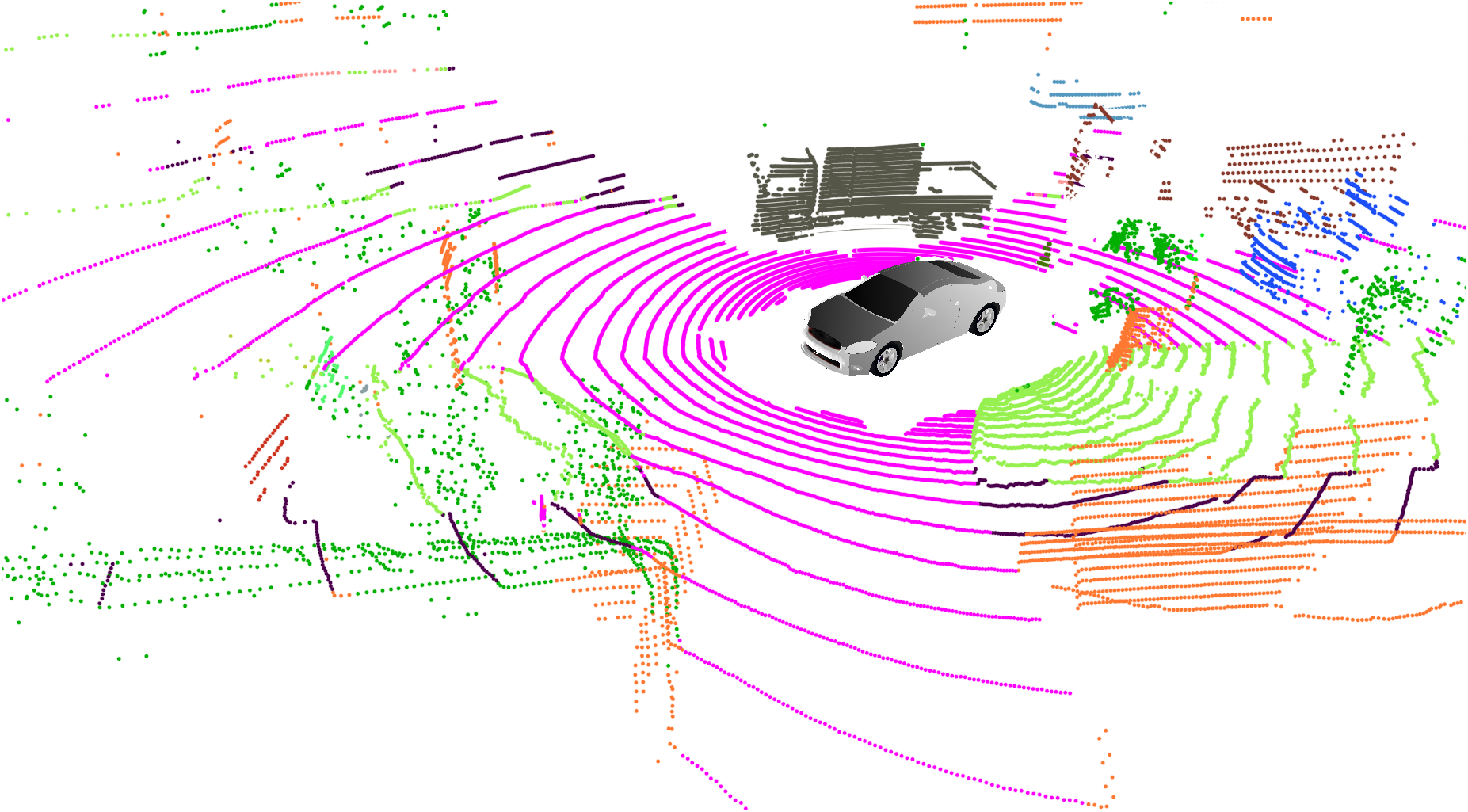}} & \raisebox{-0.4\height}{\includegraphics[width=\linewidth,frame]{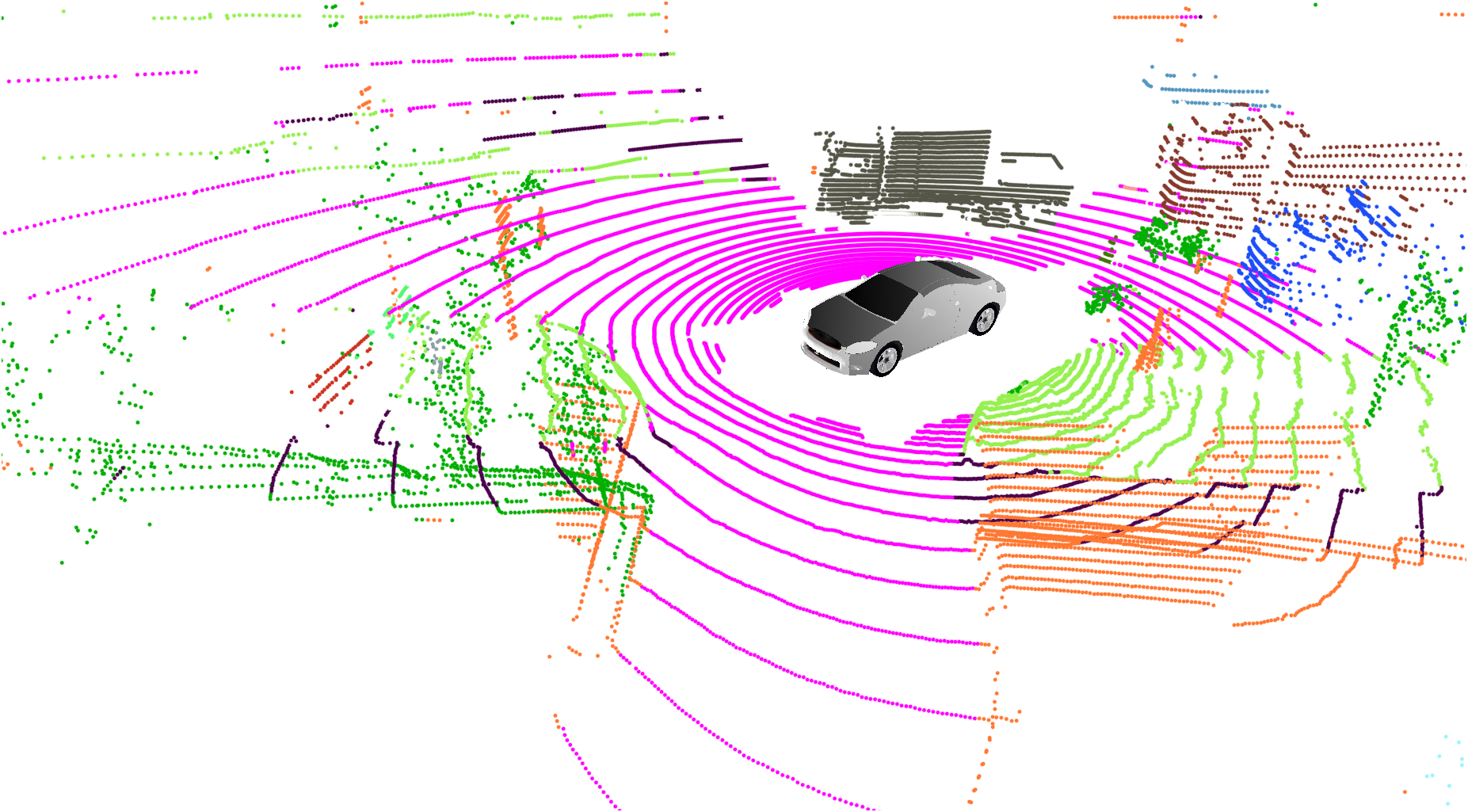}} \\
\\
\rot{(e)}  &\raisebox{-0.4\height}{\includegraphics[width=\linewidth,frame]{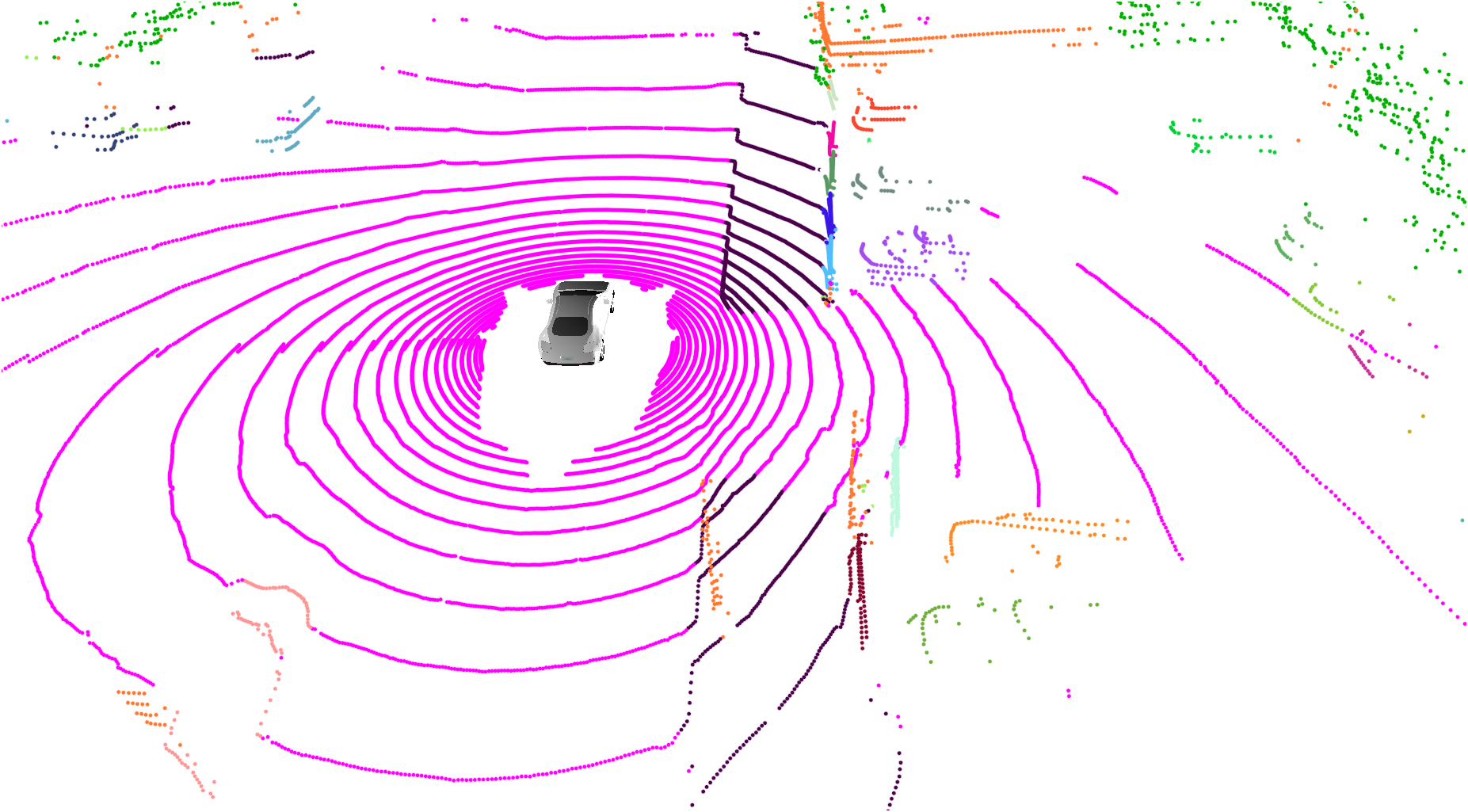}} & \raisebox{-0.4\height}{\includegraphics[width=\linewidth,frame]{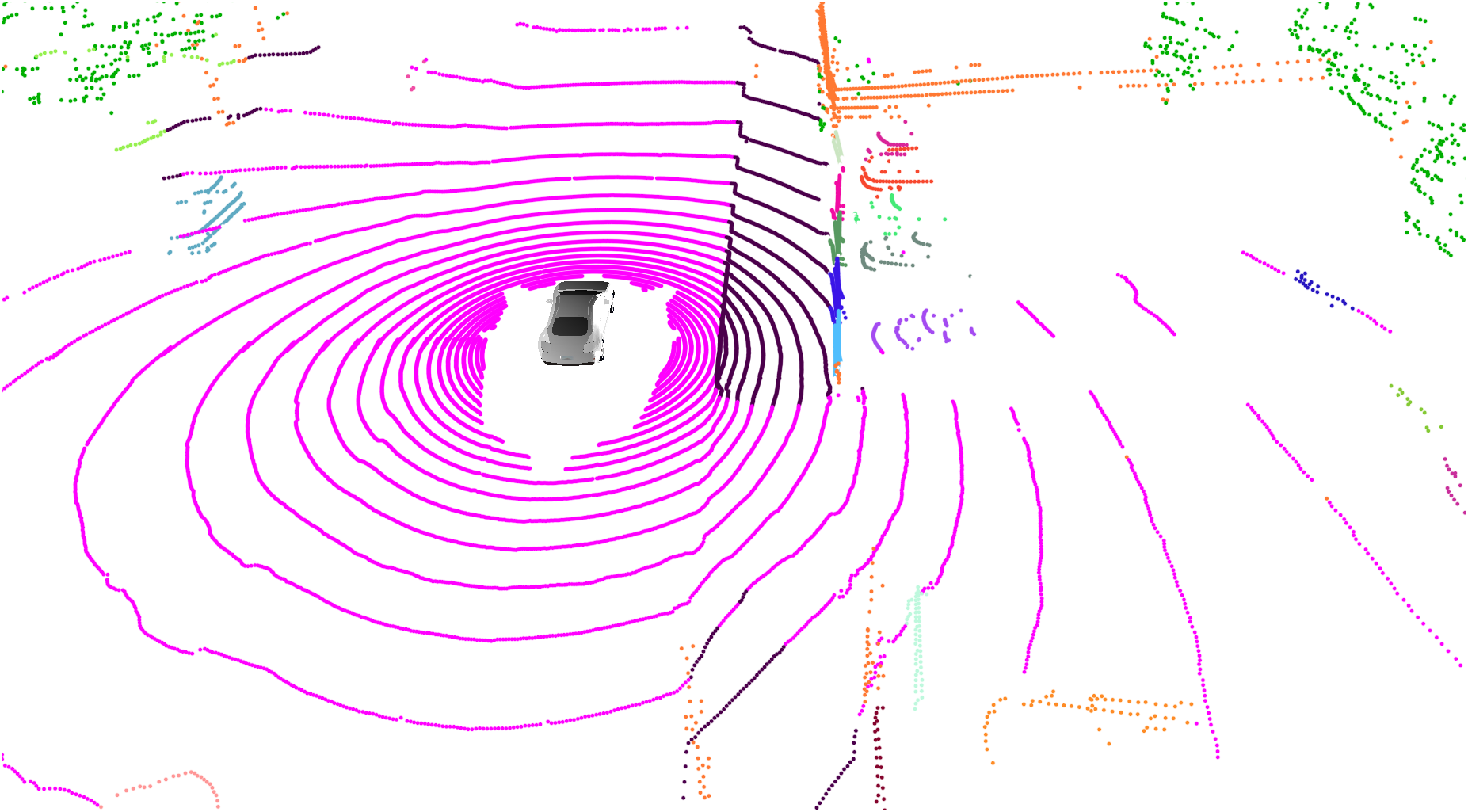}} & \raisebox{-0.4\height}{\includegraphics[width=\linewidth,frame]{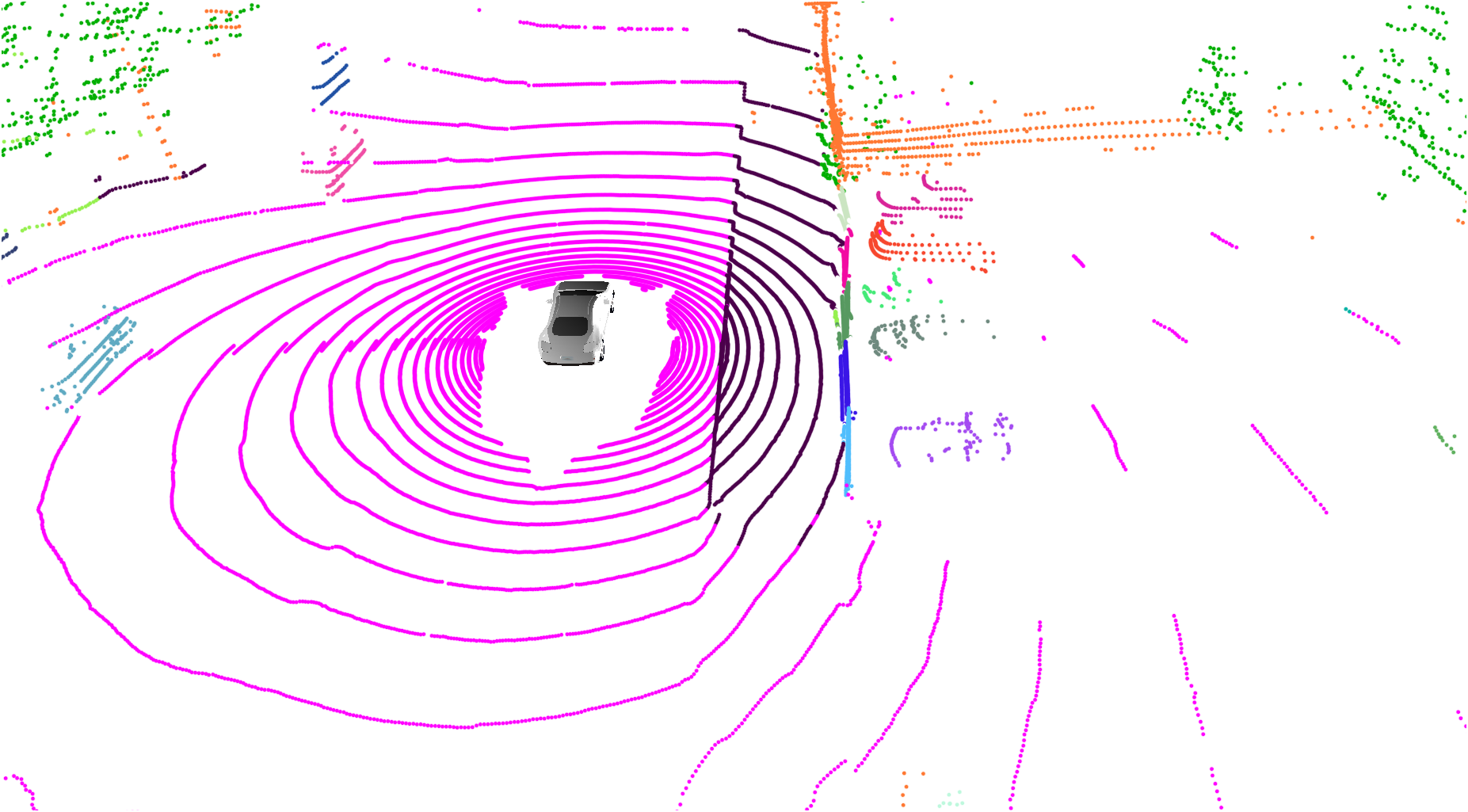}} \\
\\
\rot{(f)}  &\raisebox{-0.4\height}{\includegraphics[width=\linewidth,frame]{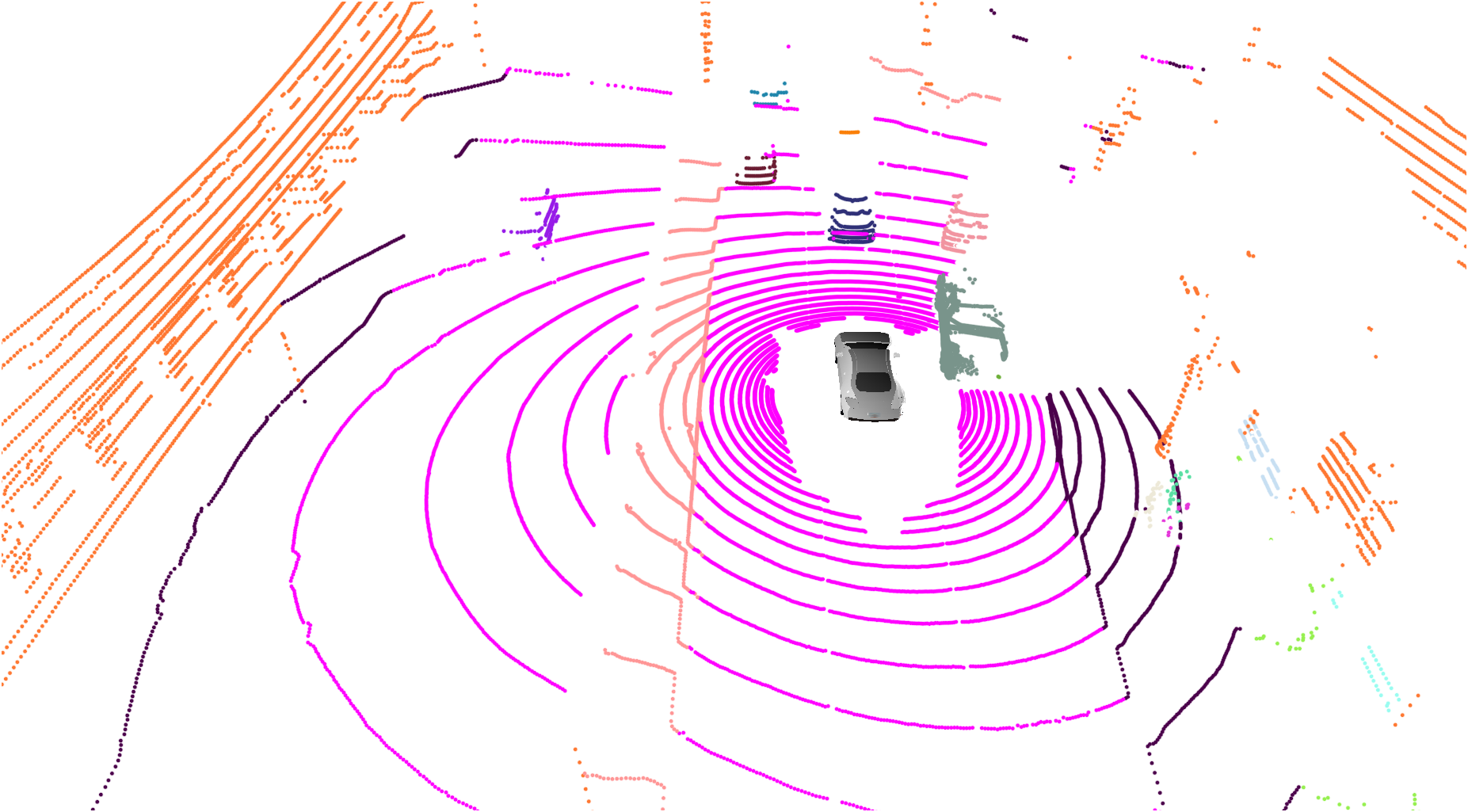}} & \raisebox{-0.4\height}{\includegraphics[width=\linewidth,frame]{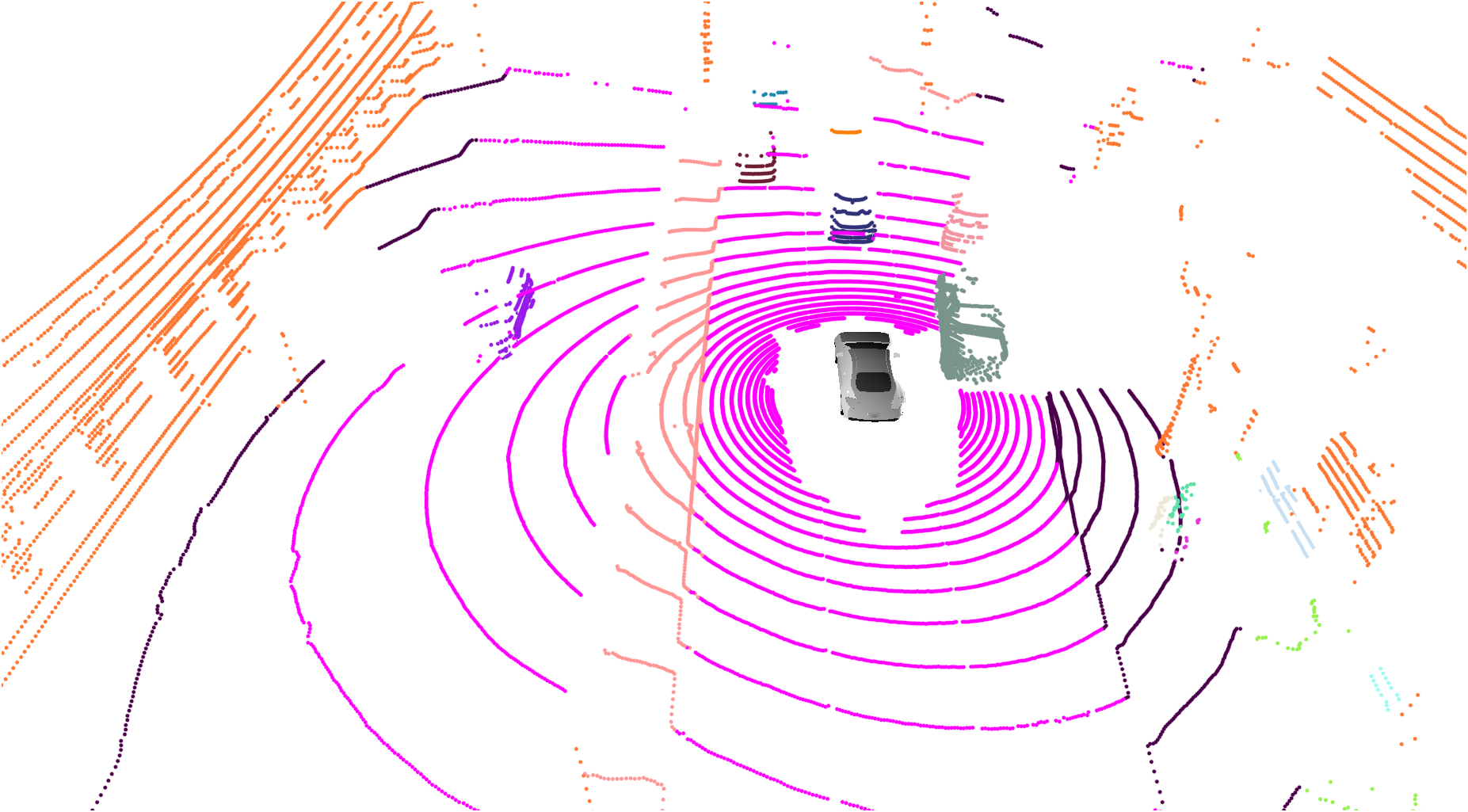}} & \raisebox{-0.4\height}{\includegraphics[width=\linewidth,frame]{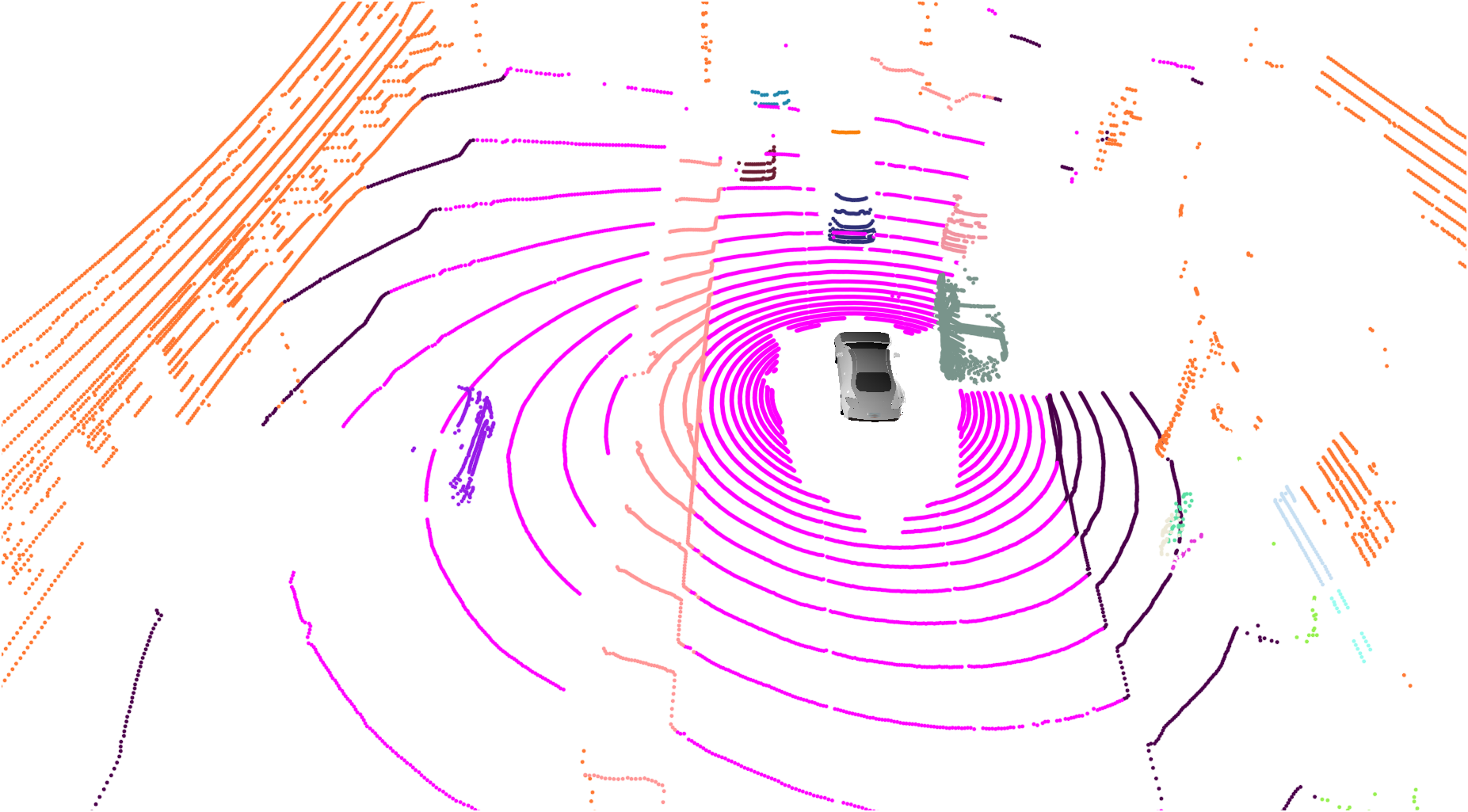}} \\
\\
\rot{(g)}  &\raisebox{-0.4\height}{\includegraphics[width=\linewidth,frame]{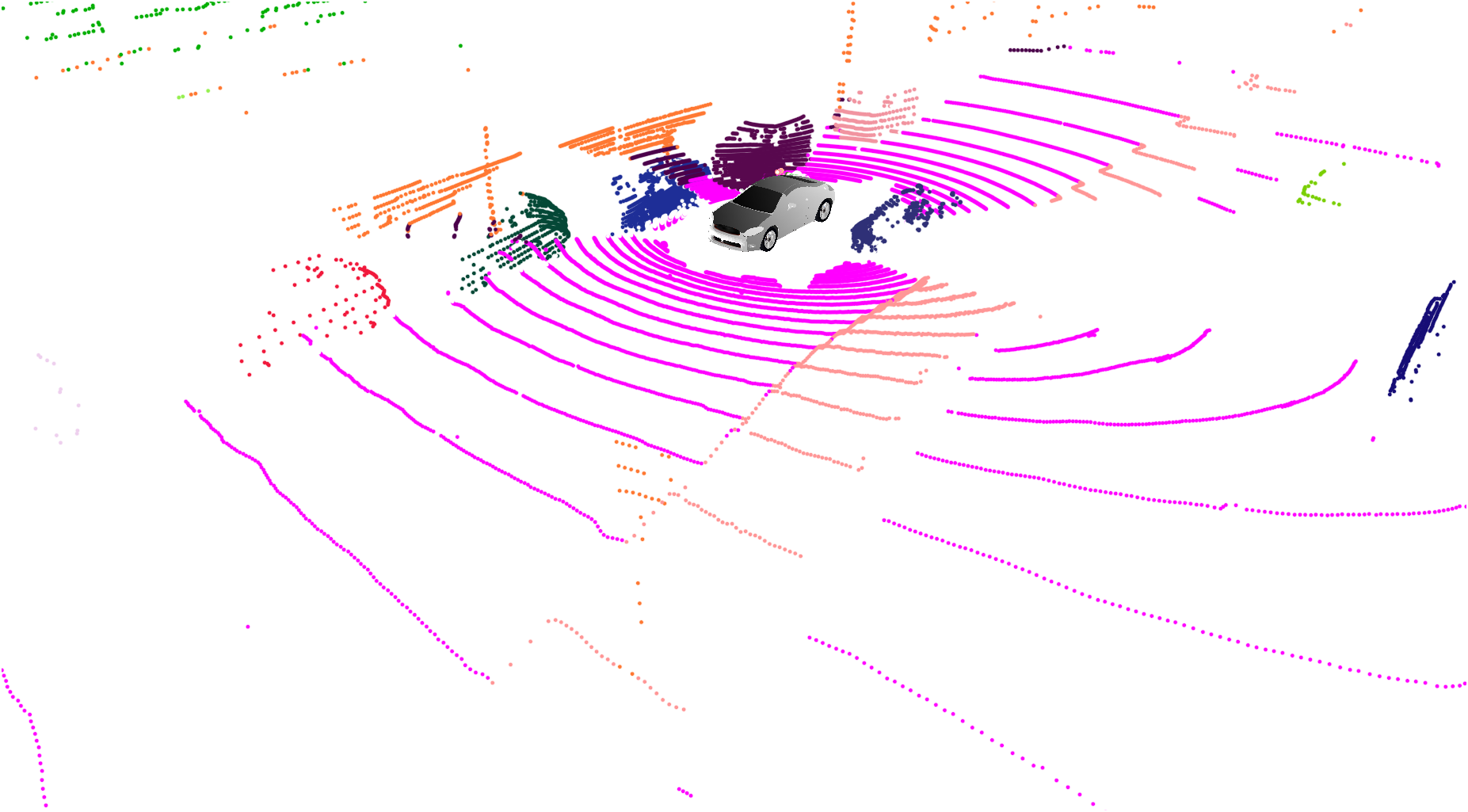}} & \raisebox{-0.4\height}{\includegraphics[width=\linewidth,frame]{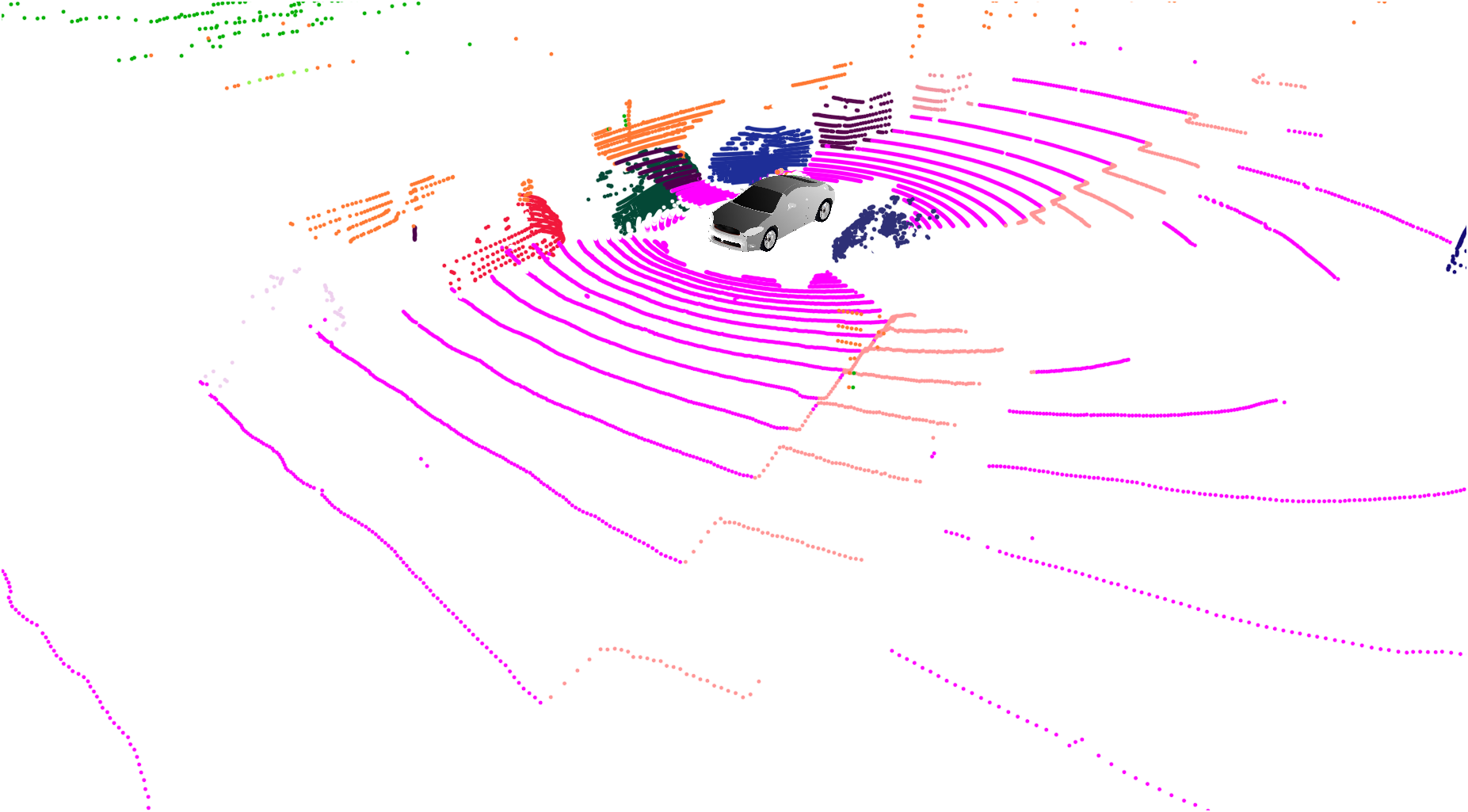}} & \raisebox{-0.4\height}{\includegraphics[width=\linewidth,frame]{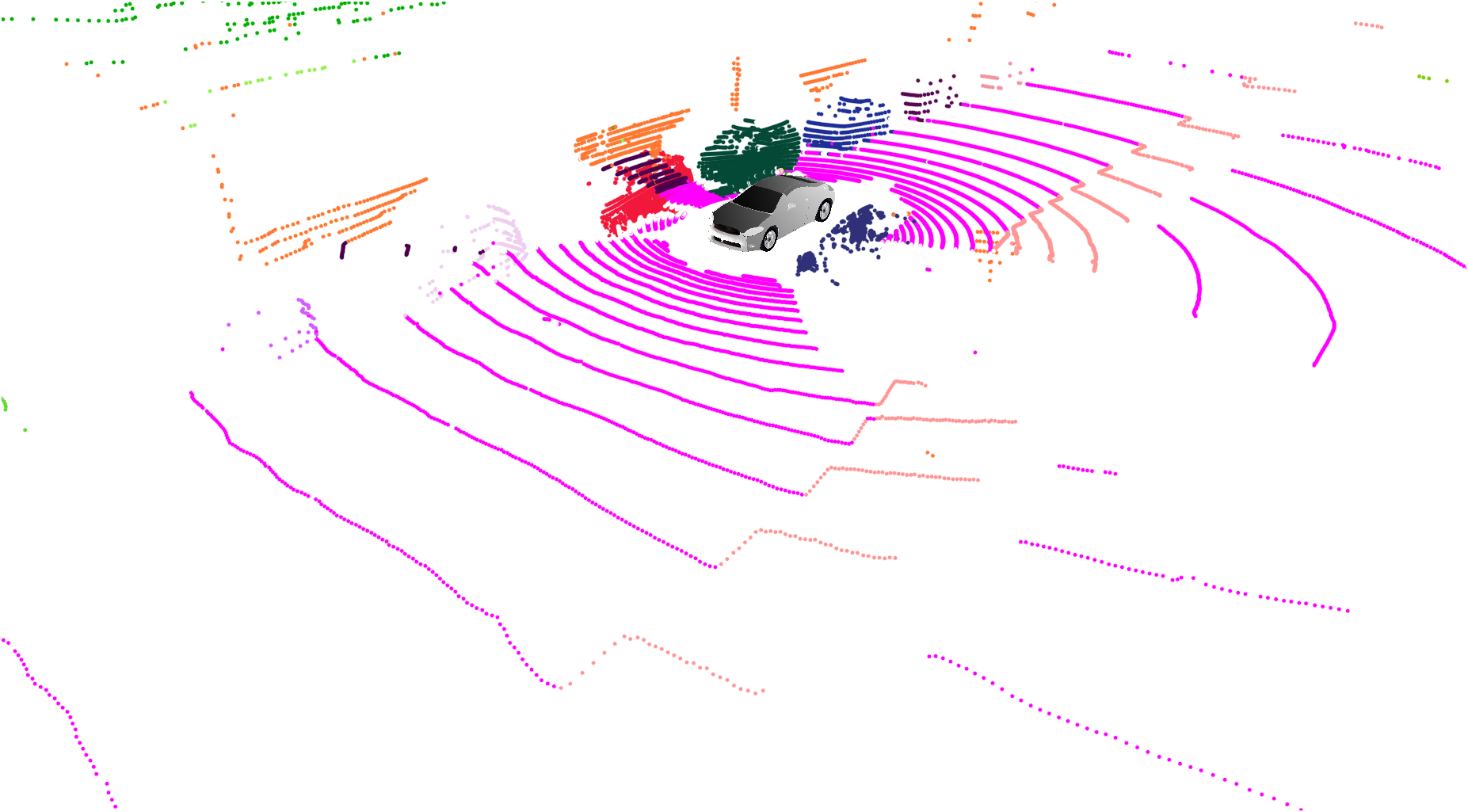}} \\
\end{tabular}}
\caption{Visualization of panoptic tracking predictions from our proposed EfficientLPT architecture on the Panoptic nuScenes dataset. Best viewed at $\times4$ zoom.}
\label{fig:visual_ablation}
\end{figure*}

In this section, we report results comparing the performance of our proposed \mbox{EfficientLPT} architecture against the baselines in the Panoptic nuScenes~\cite{fong2021panoptic} benchmark. We report the PAT, PQ, TQ, PTQ~\cite{hurtado2020mopt}, and LSTQ~\cite{aygun20214d} scores. Other metric values can be accessed via the corresponding benchmark server. \tabref{tab:resultstab} presents the benchmarking results. Among the baselines, the EfficientLPS $+$ Kalman Filter model achieves the highest PAT score of $67.1\%$. This method focuses on first predicting per scan panoptic segmentation labels and then associating the instances temporally through their state estimation. However, our proposed \mbox{EfficientLPT} outperforms all the baselines in all the metrics with a PAT score of $70.4\%$. These improvements can be attributed to the two aspects of our networks. First, the instance head has better \textit{thing} detection and segmentation quality due to the cascaded refinement stage in HTC as well as has improved capability to differentiate between foreground and background because of the feedback from the semantic head. Second, since our network treats accumulated LiDAR scans as input, the classes with fewer points in a single scan have a more defined representation in the accumulated scan. This enables improved segmentation of all the classes which in turn results in highly consistent local panoptic tracking outputs leading to an overall improved PAT score. \figref{fig:efficient_lpt} shows qualitative results of our proposed architecture. We observe that EfficientLPT consistently keeps track of all moving and non-moving objects in all of the examples. 

\section{Conclusions}
\label{sec:conclusion}

In this report, we presented our EfficientLPT architecture that achieves the
first place in the 7\textsuperscript{th} AI Driving Olympics at NeurIPS 2021 for the panoptic tracking task. The competition presents a significant challenge since it requires effectively tackling multiple tasks that were previously solved in a disjoint manner. To obtain accurate results, we require both the segmentation and tracking accuracy to be high. Consequently, the performance in the panoptic tracking challenge shows the effectiveness of our proposed EfficientLPT architecture. Additionally, the performance of our model can be considerably improved by training for more
epochs with higher batch sizes and with the inclusion of a validation set in the
training set.

{\bf Acknowledgements:} This work was funded by the Eva Mayr-Stihl Stiftung.

\bibliography{egbib}
\bibliographystyle{splncs04}

\end{document}